\documentclass[sigconf]{acmart}

\renewcommand\footnotetextcopyrightpermission[1]{} 

\usepackage[T1]{fontenc}
\usepackage[utf8]{inputenc}
\usepackage{color}
\usepackage{array}
\usepackage{multirow}
\usepackage{amssymb}
\usepackage{graphicx}
\usepackage{balance}

\makeatletter

\newcommand{\noun}[1]{\textsc{#1}}
\providecommand{\tabularnewline}{\\}

\usepackage{latexsym}

\usepackage{CJK}
\usepackage{fourier-orns}

\usepackage{amssymb}
\usepackage{pifont}
\newcommand{\cmark}{\ding{51}}
\newcommand{\xmark}{\ding{55}}

\usepackage[small,compact]{titlesec}

\@ifundefined{showcaptionsetup}{}{%
 \PassOptionsToPackage{caption=false}{subfig}}
\usepackage{subfig}
\makeatother

\setcopyright{none}

\begin{document}

\title[Logician: A Unified Neural Approach for Open-Domain Information
Extraction ]{Logician: A Unified End-to-End Neural Approach for Open-Domain
Information Extraction }

\author{Mingming Sun}
\email{sunmingming01@baidu.com}
\affiliation{Cognitive Computing Lab (CCL) Baidu Research, Beijing, China}
\author{Xu Li}
\email{lixu13@baidu.com}
\affiliation{Cognitive Computing Lab (CCL) Baidu Research, Beijing, China}
\author{Xin Wang}
\email{wangxin60@baidu.com}
\affiliation{Cognitive Computing Lab (CCL) Baidu Research, Beijing, China}
\author{Miao Fan}
\email{fanmiao@baidu.com}
\affiliation{Cognitive Computing Lab (CCL) Baidu Research, Beijing, China}
\author{Yue Feng}
\email{fengyue04@baidu.com}
\affiliation{Cognitive Computing Lab (CCL) Baidu Research, Beijing, China}
\author{Ping Li}
\email{liping11@baidu.com}
\affiliation{Cognitive Computing Lab (CCL) Baidu Research, Bellevue, WA, USA}

\begin{abstract}

In\footnote{This paper was initially submitted to EMNLP 2017. The authors sincerely thank the helpful comments from the Program Committee.} this paper, we consider the problem of open information extraction (OIE)
for extracting entity and relation level intermediate structures from sentences
in open-domain. We focus on four types of valuable intermediate structures
(Relation, Attribute, Description, and Concept), and propose a unified
knowledge expression form, SAOKE, to express them. We publicly release
a data set which contains more than forty thousand sentences and the corresponding
facts in the SAOKE format labeled by crowd-sourcing. To our knowledge,
this is the largest publicly available human labeled data set for open
information extraction tasks. Using this labeled SAOKE data set, we
train an end-to-end neural model using the sequence-to-sequence paradigm,
called Logician, to transform sentences into facts. For each sentence,
different to existing algorithms which generally focus on extracting
each single fact without concerning other possible facts,  Logician
performs a global optimization over all possible involved facts, in
which facts not only compete with each other to attract the attention of words,
but also cooperate to share words. An experimental study on various
types of open domain relation extraction tasks reveals the consistent
superiority of Logician to other states-of-the-art algorithms. The experiments
verify the reasonableness of SAOKE format, the valuableness of SAOKE
data set, the effectiveness of the proposed Logician model, and
the feasibility of the methodology to apply end-to-end learning paradigm
on supervised data sets for the challenging tasks of open information extraction.

\vspace{0.06in}

\noindent 
\textbf{SAOKE}:\hspace{0.2in} 
\url{https://ai.baidu.com/broad/subordinate?dataset=saoke}

\end{abstract}

\keywords{Knowledge expression; open information extraction; end-to-end learning; sequence-to-sequence learning; deep learning}

\maketitle

\section{Introduction}

Semantic applications typically work on the basis of intermediate structures
derived from sentences. Traditional word-level intermediate structures,
such as POS-tags, dependency trees and semantic role labels, have been
widely applied. Recently, entity and relation level intermediate structures
attract increasingly more attentions.

\vspace{0.06in}

In general, knowledge based applications
require entity and relation level information. For instance, in~\cite{Riedel2013},
the lexicalized dependency path between two entity mentions was taken
as the surface pattern facts. In distant supervision~\cite{Mintz2009},
the word sequence and dependency path between two entity mentions
were taken as evidence of certain relation. In Probase~\cite{Wu2012},
candidates of taxonomies were extracted by Hearst patterns~\cite{Hearst1992}.
The surface patterns of relations extracted by Open Information Extraction
(OIE) systems~\cite{Banko2007a,Etzioni2011,Schmitz2012,Fu2010,Fader2011}
worked as the source of question answering systems ~\cite{Khot2017,Fader2014}.
In addition, entity and relation level intermediate structures
have been proven effective in many other tasks such as text summarization~\cite{Christensen2013,Christensen2014,Mausam2016},
text comprehension, word similarity, word analogy~\cite{Stanovsky2015}, and more.

\vspace{0.06in}

The task of entity/relation level mediate structure extraction
studies how facts about entities and relations are expressed by natural
language in sentences, and then expresses these facts in an intermediate (and convenient)
format.  Although entity/relation level intermediate structures have been
utilized in many applications, the study of learning these structures is still in an early stage.

Firstly, the problem of extracting different types of entity/relation level intermediate
structures has not been considered in a unified fashion. Applications
generally need to construct their own handcrafted heuristics to extract
required entity/relation level intermediate structures, rather than
consulting a commonly available NLP component, as they do for word
level intermediate structures. Open IE-v4 system (http://knowitall.github.io/openie/)
attempted to build such components by developing two sub-systems, with each   extracting one type of intermediate structures, i.e., SRLIE~\cite{Christensen2011}
for verb based relations, and ReNoun~\cite{Pal2016,Yahya2014} for
nominal attributes. However, important information about descriptive
tags for entities and concept-instance relations between entities
were not considered.

Secondly, existing solutions to the task either used pattern matching technique~\cite{Wu2012,Banko2007a,Schmitz2012,Fu2010},
or were trained in a self-supervised manner on the data set automatically generated by heuristic
patterns or info-box matching~\cite{Fu2010,Banko2007a,Fader2011}.
It is well-understood that pattern matching typically  does not generalize well and the automatically generated samples may contain  lots of noises.


This paper aims at tackling some of the well-known challenging problems in OIE systems, in a supervised end-to-end deep learning paradigm. Our contribution can be summarized as three major components: \textbf{SAOKE format}, \textbf{SAOKE data set}, and \textbf{Logician}.

\vspace{0.06in}

\textbf{Symbol Aided Open Knowledge Expression (SAOKE)} is a knowledge
expression form with several desirable properties: (i) SAOKE is literally honest and open-domain.
Following the philosophy of OIE systems, SAOKE uses words in the
original sentence to express knowledge.   (ii) SAOKE provides a unified view over four common types of knowledge: {\em relation}, {\em attribute},
{\em description} and {\em concept}.  (iii) SAOKE is an accurate expression.  With the aid of symbolic
system, SAOKE is able to accurately express facts with separated relation
phrases, missing information, hidden information, etc.

\textbf{SAOKE Data Set} is a human annotated data set containing 48,248
 Chinese sentences and corresponding facts in the SAOKE
form. We publish the data set for research purpose. To the best of our
knowledge, this is the largest publicly available human annotated data
set for open-domain information extraction tasks.

\textbf{Logician} is a supervised end-to-end neural learning algorithm
which transforms natural language sentences into facts in the SAOKE form.
Logician is trained under the attention-based sequence-to-sequence
paradigm, with three mechanisms: restricted copy mechanism to ensure literally
honestness, coverage mechanism to alleviate the under extraction and
over extraction problem, and gated dependency attention mechanism
to incorporate dependency information. Experimental results on four
types of open information extraction tasks reveal the superiority of
the Logician algorithm.

\vspace{0.06in}

Our work will demonstrate that SAOKE format is suitable for expressing
various types of knowledge and is friendly to end-to-end learning algorithms. Particularly, we will focus on showing that
the supervised end-to-end learning is promising  for OIE tasks, to extract entity and
relation level intermediate structures.

The rest of this paper is organized as follows. Section~\ref{sec:SAOKE}
presents the details of SAOKE. Section~\ref{sec:SAOKE-Data-Set}
describes the human labeled SAOKE data set. Section~\ref{sec:Logician}
describes the Logician algorithm and Section~\ref{sec:Experimental-Results}
evaluates the Logician algorithm and compares its performance with
the state-of-the-art algorithms on four OIE tasks. Section~\ref{sec:Related-work}
discusses the related work and Section~\ref{sec:Conclusion} concludes
the paper.

\section{SAOKE Format: \ Symbol Aided Open Knowledge Expression\label{sec:SAOKE}}

\begin{table*}[t]
\caption{Expected facts of an example sentence.\label{tab:Expected-facts-example}\vspace{0in} }
{\small
\begin{tabular}{|>{\raggedright}p{0.08\textwidth}|>{\raggedright}p{0.37\textwidth}|>{\raggedright}p{0.47\textwidth}|}
\hline
 & Chinese & English Translation\tabularnewline
\hline
\hline
Sentence & \begin{CJK}{UTF8}{gbsn}
李白(701年－762年), 深受庄子思想影响，爽朗大方，
爱饮酒作诗，喜交友，
代表作有《望庐山瀑布》等著名诗歌。\end{CJK} & Li Bai (701 - 762), with masterpieces of famous poetries such as
\textquotedbl{}Watching the Lushan Waterfall\textquotedbl{}, was deeply
influenced by Zhuangzi's thought, hearty and generous, loved to drink
and write poetry, and liked to make friends.\tabularnewline
\hline
\hline
Relations & \begin{CJK}{UTF8}{gbsn}
(李白,  深受X影响, 庄子思想) \newline
(李白,  爱,  [饮酒|作诗])
(李白,  喜,  交友)  \end{CJK} & (Li Bai, deeply influenced by, Zhuangzi's thought)

(Li Bai, loved to, {[}drink| write poetry{]}) (Li Bai, liked to,
make friends)\tabularnewline
\hline
Attribute & \begin{CJK}{UTF8}{gbsn}
(李白, BIRTH, 701年)
(李白,  DEATH,  762年) \newline
(李白,  代表作,  《望庐山瀑布》)
\end{CJK} & (Li Bai, BIRTH, 701)(Li Bai, DEATH, 762)

(Li Bai, masterpiece, \textquotedbl{}Watching the Lushan Waterfall\textquotedbl{})\tabularnewline
\hline
Description & \begin{CJK}{UTF8}{gbsn}
(李白, DESC, 爽朗大方)
\end{CJK} & (Li Bai, DESC, hearty and generous)\tabularnewline
\hline
Concept & \begin{CJK}{UTF8}{gbsn}
(《望庐山瀑布》, ISA, 著名诗歌)
\end{CJK} & (\textquotedbl{}Watching the Lushan Waterfall\textquotedbl{}, ISA,
famous poetry)\tabularnewline
\hline
\end{tabular}
}

\end{table*}

When reading a sentence in natural language, humans are able to recognize the
facts involved in the sentence and  accurately express them. In this paper, Symbolic Aided Open Knowledge Expression (SAOKE) is proposed
as the form for honestly recording these facts. SAOKE expresses the primary
information of sentences in n-ary tuples $(subject,predicate,object_{1},\cdots,object_{N})$,
and (in this paper) neglects some auxiliary information. In the design of SAOKE,
we take four requirements into consideration: completeness, accurateness,
atomicity and compactness.

\subsection{Completeness}
\label{subsec:SAOKE_Complete}

After having analyzed a large number of sentences, we observe that the majority of
facts  can be classified into the following classes:
\begin{enumerate}
\item
\textbf{\noun{Relation:}}\noun{ }Verb/preposition based n-ary relations
between entity mentions~\cite{Christensen2011,Schmitz2012};
\item
\textbf{\noun{Attribute:}}Nominal attributes for entity mentions~\cite{Pal2016,Yahya2014};
\item
\textbf{\noun{Description:}} Descriptive phrases of entity mentions~\cite{Chakrabarti2011};
\item
\textbf{\noun{Concept:}}\noun{ }Hyponymy and synonym relations among
concepts and instances~\cite{VeredShwartz2016a}.
\end{enumerate}

SAOKE is designed to express all these four types of facts. Table~\ref{tab:Expected-facts-example}
presents an example  sentence and the involved facts of these four
classes in the SAOKE form. We should mention  that the sentences and facts in English
are directly translated from the corresponding Chinese sentences and
facts, and the facts in English may not be the desired outputs of
OIE algorithms for those English sentences due to the differences between
Chinese and English languages.

\subsection{Accurateness}

SAOKE adopts the ideology of ``literally honest''. That is, as much as possible, it uses
the words in the original sentences to express the facts. SAOKE follows the philosophy of OIE systems to express various
relations without relying on any predefined schema system. There are, however, exceptional situations which are beyond the expression ability of this format. Extra symbols will be introduced to handle these situations, which are explained as follows.

\vspace{0.06in}

\textbf{Separated relation phrase:}\ In some languages such as Chinese,
relation phrases may be divided into several parts residing in discontinued
locations of the sentences. To accurately express these relation phrases,
we add placeholders ($X$,$Y$,$Z$, etc) to build continuous and
complete expressions. \begin{CJK}{UTF8}{gbsn} ``深受X影响'' \end{CJK}
(``deeply influenced by X'' in English) in the example of Table~\ref{tab:Expected-facts-example}
is an instance of relation phrase after such processing.

\textbf{Abbreviated expression:}\ We explicitly express the information
in abbreviated expressions by introducing symbolic predicates. For
example, the expression of ``Person (birth date - death date)'' is
transformed into facts: (Person, BIRTH, birth date) (Person, DEATH,
death date), and the synonym fact involved in ``NBA (National Basketball
Association)'' is expressed in the form of (NBA, = , National Basketball
Association) .

\textbf{Hidden information:}\ Description of an entity and hyponymy
relation between entities are in general expressed implicitly in sentences,
and are expressed by symbolic predicates ``DESC'' and ``ISA'' respectively,
as  in Table~\ref{tab:Expected-facts-example}. Another source
of hidden information is the address expression. For example, \begin{CJK}{UTF8}{gbsn} ``法国巴黎'' \end{CJK}
(``Paris, France'' in English) implies the fact \begin{CJK}{UTF8}{gbsn} (巴黎, LOC, 法国)  \end{CJK}
((Paris, LOC, France) in English), where the symbol ``LOC'' means
``location''.

\textbf{Missing information:}\ A sentence may not tell us the exact
relation between two entities, or the exact subject/objects of a relation,
which are required to be inferred from the context. We use placeholders
like ``$X,Y,Z$'' to denote the missing subjects/objects, and ``$P$''
to denote the missing predicates.

\subsection{Atomicity}

Atomicity is introduced to eliminate the ambiguity of knowledge expressions.
In SAOKE format, each fact is required to be atomic, which means that:
(i) it is self-contained for an accurate expression; (ii) it cannot be decomposed
into multiple valid facts. We provide examples in Table~\ref{tab:Example-actom}
to help understand these two criteria.

Note that the second criterion implies that any logical
connections (including nested expressions) between facts are neglected
(e.g., the third case in Table~\ref{tab:Example-actom} ). This problem
of expression relations between facts will be considered in the future
version of SAOKE.

\begin{table*}
\caption{Example sentence and corresponding wrong/correct facts.\vspace{0in} \label{tab:Example-actom}}
{\small
\begin{tabular}{|>{\raggedright}p{0.05\textwidth}|>{\raggedright}p{0.24\textwidth}|>{\raggedright}p{0.3\textwidth}|>{\raggedright}p{0.3\textwidth}|}
\hline
 & Sentence & Wrong Facts & Correct Facts\tabularnewline
\hline
\hline
Chinese & \begin{CJK}{UTF8}{gbsn}山东的GDP高于所有西部的省份。\end{CJK} & \begin{CJK}{UTF8}{gbsn} (山东的GDP, 高于, 所有省份) (所有省份, DESC, 西部的)\end{CJK} & \begin{CJK}{UTF8}{gbsn} (山东的GDP, 高于, 西部的所有省份)\end{CJK}\tabularnewline
\hline
English & Shandong's GDP is higher than in all western provinces. & (Shandong's GDP, is higher than, all provinces) (all provinces, DESC,
western)  & (Shandong's GDP, is higher than, all western provinces) \tabularnewline
\hline
\hline
Chinese & \begin{CJK}{UTF8}{gbsn}李白游览了雄奇灵秀的泰山。\end{CJK} & \begin{CJK}{UTF8}{gbsn} (李白, 游览, 雄奇灵秀的泰山) \end{CJK} & \begin{CJK}{UTF8}{gbsn} (李白, 游览, 泰山) (泰山, DESC, 雄奇灵秀)\end{CJK}\tabularnewline
\hline
English & Li Bai visited the magnificent Mount Tai. & (Li Bai, visited, the magnificent Mount Tai) & (Li Bai, visited, the Mount Tai)(the Mount Tai, DESC, magnificent)\tabularnewline
\hline
\hline
Chinese & \begin{CJK}{UTF8}{gbsn}在美国的帮助下，英国抵挡住了德国的进攻。\end{CJK} & \begin{CJK}{UTF8}{gbsn} (英国, 在X的帮助下抵挡住了Y的进攻, 美国, 德国)\end{CJK} & \begin{CJK}{UTF8}{gbsn} (美国, 帮助, 英国)(英国, 抵挡住了X的进攻, 德国)\end{CJK}\tabularnewline
\hline
English & With the help of the US, the British resisted the attack from German. & (the British, resisted the attack from X with the help of Y, German,
the US) & (the US, helped, the British)( the British, resisted the attack from,
German)\tabularnewline
\hline
\end{tabular}}

\end{table*}

\subsection{Compactness}

Natural language may express several facts in a compact form. For
example, in a sentence \begin{CJK}{UTF8}{gbsn} ``李白爱饮酒作诗''\end{CJK}
(``Li Bai loved to drink and write poetry'' in English ), according
to atomicity, two facts should be extracted: \begin{CJK}{UTF8}{gbsn} (李白,  爱,  饮酒)(李白,  爱,  作诗)\end{CJK}
( (Li Bai, loved to, drink)(Li Bai, loved to, write poetry) in English
). In this situation, SAOKE  adopts a compact expression to merge
these two facts into one expression: \begin{CJK}{UTF8}{gbsn} (李白,  爱,  [饮酒|作诗])\end{CJK}
( (Li Bai, loved to, {[}drink| write poetry{]}) in English ).

The compactness of expressions is introduced to fulfill, but not to violate
the rule of ``literally honest''. SAOKE does not allow merging facts
if facts are not expressed compactly in original sentences. By
this means, the differences between the sentences and the corresponding knowledge expressions
are reduced, which  may help reduce the complexity of learning from
data in SAOKE form.

With the above designs, SAOKE is able to express various kinds of facts,
with each historically considered by different open information extraction
algorithms, for example, verb based relations in SRLIE~\cite{Christensen2011}
and nominal attributes in ReNoun~\cite{Pal2016,Yahya2014}, descriptive
phrases for entities in EntityTagger~\cite{Chakrabarti2011}, and
hypernyms in HypeNet~\cite{VeredShwartz2016a}. SAOKE introduces the
atomicity to eliminate the ambiguity of knowledge expressions, and
achieves better accuracy and compactness with the aid of the symbolic
expressions.

\section{SAOKE Data Set\label{sec:SAOKE-Data-Set}}

We randomly collect sentences from Baidu Baike (\url{http://baike.baidu.com}), and send those sentences to a
crowd sourcing company to label the involved facts. The workers are trained with
labeling examples and tested with exams. Then the workers with high exam scores are
asked to read and understand the facts  in the sentences,
and express the facts in the SAOKE format. During the procedure, one sentence is only labeled by one worker.
Finally, more than forty thousand sentences with about one hundred thousand facts are returned to us.
The manual evaluation
results on 100 randomly selected sentences show that
the fact level precision and recall is 89.5\% and
92.2\% respectively.
Table~\ref{tab:Components-of-SAOKE} shows the proportions of
four types of facts (described in Section \ref{subsec:SAOKE_Complete})
contained in the data set. Note that the facts
with missing predicates represented by ``P'' are classified into ``Unknown''.
We publicize the data
set at \url{https://ai.baidu.com/broad/subordinate?dataset=saoke}.

\begin{table}[h!]
\caption{Ratios of facts of each type in SAOKE data set.\vspace{-0in}\label{tab:Components-of-SAOKE}}
{\small\begin{tabular}{cccccc}
\hline
 & Relation & Attribute & Description & Concept & Unknown\\
\hline
Ratio & 76.02\% & 7.25\% & 9.89\% & 3.64\% & 3.20\%\\
\hline
\end{tabular}
}
\end{table}

Prior to the SAOKE data set, an annotated data set for OIE tasks with
3,200 sentences in 2 domains was released in~\cite{Stanovsky2016}
to evaluate OIE algorithms, in which the data set was said~\cite{Stanovsky2016} ``13 times
larger than the previous largest annotated Open IE corpus''.
The SAOKE data set is 16 times larger than the data set in~\cite{Stanovsky2016}. To the best of our knowledge,
SAOKE data set is the largest publicly available human labeled data
set for OIE tasks. Furthermore, the data set released in~\cite{Stanovsky2016}
was generated from a QA-SRL data set~\cite{He2015a}, which indicates
that the data set only contains facts that can be discovered by SRL
(Semantic Role Labeling) algorithms, and thus is biased, whereas the
SAOKE data set is not biased to an algorithm. Finally, the SAOKE
data set contains sentences and facts from a large number of domains.

\section{Logician\label{sec:Logician}}

Given a sentence $S$ and a set of expected facts (with all the possible types of facts)
$\mathbb{F}=\{F_{1},\cdots,F_{n}\}$
in SAOKE form, we join all the facts in the order that annotators
wrote them into a char sequence $F$ as the expected output.  We
build Logician under the attention-based sequence-to-sequence
learning paradigm, to transform $S$ into $F$, together with the restricted
copy mechanism, the coverage mechanism and the gated dependency mechanism.
\subsection{Attention based Sequence-to-sequence Learning }

The attention-based sequence-to-sequence learning~\cite{DzmitryBahdana2014}
have been successfully applied to the task of generating text and patterns.
Given an input sentence $S=[w_{1}^{S},\cdots,w_{N_{S}}^{S}]$, the
target sequence $F=[w_{1}^{F},\cdots,w_{N_{F}}^{F}]$ and a vocabulary
$V$ (including the symbols introduced in Section~\ref{sec:SAOKE}
and the OOV (out of vocabulary) tag ) with size $N_{v}$, the words
$w_{i}^{S}$ and $w_{j}^{F}$ can be represented as one-hot vectors
$v_{i}^{S}$ and $v_{j}^{F}$ with dimension $N_{v}$, and transformed
into $N_{e}$-dimensional distributed representation vectors by an
embedding transform $x_{i}=Ev_{i}^{S}$ and $y_{j}=Ev_{j}^{F}$ respectively,
where $E\in\mathbb{R}^{(N_{e}\times N_{v})}$. Then the sequence of
$\{x_{i}\}_{i=1}^{N_{s}}$ is transformed into a sequence of $N_{h}$-dimensional
hidden states $H^{S}=[h_{1}^{S},\cdots,h_{N_{S}}^{S}]$ using bi-directional
GRU (Gated Recurrent Units) network~\cite{Cho2014}, and the sequence
of $\{y_{j}\}_{j=1}^{N_{F}}$ is transformed into a sequence of $N_{h}$-dimensional
hidden states $H^{F}=[h_{1}^{F},\cdots,h_{N_{F}}^{F}]$ using GRU
network.

For each position $t$ in the target sequence, the decoder learns
a dynamic context vector $c_{t}$ to focus attention on specific location
$l$ in the input hidden states $H^{S}$, then computes the probability
of generated words by $p(w_{t}^{F}|\{w_{1}^{F},\cdots,w_{t-1}^{F}\},c_{t})=g(h_{t-1}^{F},s_{t},c_{t})$,
where $s_{t}$ is the hidden state of the GRU decoder, $g$ is the
word selection model (details could be found in~\cite{DzmitryBahdana2014}),
and $c_{t}$ is computed as $c_{t}=\sum_{j=1}^{N_{S}}\alpha_{tj}h_{j},$where
$\alpha_{tj}=\frac{\exp(e_{tj})}{\sum_{k=1}^{N_{S}}\exp(e_{tk})}$
and $e_{tj}=a(s_{t-1},h_{j}^{S})=v_{a}^{T}\tanh(W_{a}s_{t-1}+U_{a}h_{j}^{S})$
is the alignment model to measure the strength of focus on the $j$-th
location. $W_{a}\in\mathbb{R}^{(N_{h}\times N_{h})}$,$U_{a}\in\mathbb{R}^{(N_{h}\times N_{h})}$,
and $v_{a}\in\mathbb{R}^{N_{h}}$are weight matrices.

\subsection{Restricted Copy Mechanism}

The word selection model employed in~\cite{DzmitryBahdana2014} selects
words from the whole vocabulary $V$, which evidently violates the
``literal honest'' requirement of SAOKE. We propose a restricted version
of copy mechanism~\cite{Gu2016} as the word selection model for
Logician:

We collect the symbols introduced in Section~\ref{sec:SAOKE} into
a keyword set $K=\{$``$ISA$'', ``$DESC$'', ``$LOC$'', ``$BIRTH$'',
``$DEATH$'', ``$=$'', ``$($'', ``)'', ``$\$$'',``$[$'', ``$]$'',
``$|$'', ``$X$'', ``$Y$'', ``$Z$'', ``$P$''$\}$ where ``$\$$''
is the separator of elements of fact tuples. ``$X$'', ``$Y$'', ``$Z$'',
``$P$'' are placeholders . When the decoder is considering generating
a word $w_{t}^{F}$, it can choose $w_{t}^{F}$ from either $S$ or
$K$.
\begin{equation}\label{eq:total_prob}
p(w_{t}^{F}|w_{t-1}^{F},s_{t},c_{t})=p_{X}(w_{t}^{F}|w_{t-1}^{F},s_{t},c_{t})+p_{K}(w_{t}^{F}|w_{t-1}^{F},s_{t},c_{t}),
\end{equation}
where $p_{X}$ is the probability of copying from $S$ and $p_{K}$
is the probability of selecting from $K$. Since $S\cap K=\phi$ and
there are no unknown words in this problem setting, we compute $p_{X}$
and $p_{K}$ in a simpler way than that in~\cite{Gu2016}, as follows:
\begin{eqnarray*}
p_{X}(w_{t}^{F}=w_{j}^{S}) & = & \frac{1}{Z}\exp(\sigma((h_{j}^{S})^{T}W_{c})s_{t}),\\
p_{K}(w_{t}^{F}=k_{i}) & = & \frac{1}{Z}\exp(v_{i}^{T}W_{o}s_{t}),
\end{eqnarray*}
where the (generic) $Z$ is the normalization term, $k_{i}$ is one of keywords,
$v_{i}$ is the one-hot indicator vector for $k_{i}$, $W_{o}\in\mathbb{R}^{(|K|\times N_{h})}$,$W_{c}\in\mathbb{R}^{(N_{h}\times N_{h})}$,
and $\sigma$ is a nonlinear activation function.

\subsection{Coverage Mechanism}

In practice, Logician may forget to extract some facts (\emph{under-extraction})
or extract the same fact many times (\emph{over-extraction}). We
incorporate the coverage mechanism~\cite{Tu2016} into Logician to alleviate
these problems. Formally, when the decoder considers generating a
word $w_{t}^{F}$, a coverage vector $m_{j}^{t}$ is introduced for
each word $w_{j}^{S}$ , and updated as follows:
\begin{eqnarray*}
m_{j}^{t} & = & \mu(m_{j}^{t-1},\alpha_{tj},h_{j}^{S},s_{t-1})=(1-z_{i})\circ m_{j}^{t-1}+z_{j}\circ\tilde{m}_{j}^{t},\\
\tilde{m}_{j}^{t} & = & \tanh(W_{h}h_{j}^{S}+u_{\alpha}\alpha_{tj}+W_{s}s_{t-1}+U_{m}[r_{i}\circ m_{j}^{t-1}]),
\end{eqnarray*}
where $\circ$ is the element-wise multiplication operator. The update
gate $z_{j}$ and the reset gate $r_{j}$ are defined as, respectively,
\begin{eqnarray*}
z_{j} & = & \sigma(W_{h}^{z}h_{j}^{S}+u_{\alpha}^{z}\alpha_{tj}+W_{s}^{z}s_{t-1}+U_{m}^{z}m_{j}^{t-1}),\\
r_{j} & = & \sigma(W_{h}^{r}h_{j}^{S}+u_{\alpha}^{r}\alpha_{tj}+W_{s}^{r}s_{t-1}+U_{m}^{r}m_{j}^{t-1}),
\end{eqnarray*}
where $\sigma$ is a logistic sigmoid function. The coverage vector
$m_{j}^{t}$ contains the information about the historical attention
focused on $w_{j}^{S}$, and is helpful for deciding whether $w_{j}^{S}$
should be extracted or not. The alignment model is updated as follows~\cite{Tu2016}:
\[
e_{tj}=a(s_{t-1},h_{j}^{S},m_{j}^{t-1})=v_{a}^{T}\tanh(W_{a}s_{t-1}+U_{a}h_{j}^{S}+V_{a}m_{j}^{t-1}),
\]
where $V_{a}\in\mathbb{R}^{(N_{h}\times N_{h})}$.

\subsection{Gated Dependency Attention}

The semantic relationship between candidate words and the previously
decoded word is valuable to guide the decoder to select the correct
word. We introduce the gated dependency attention mechanism to utilize
such guidance.

For a sentence $S$, we extract the dependency tree using NLP tools
such as CoreNLP~\cite{Manning2014} for English and LTP~\cite{Che2010}
for Chinese, and convert the tree into a graph by adding reversed
edges with a revised labels (for example, adding $w_{j}^{S}\xrightarrow{-SBV}w_{i}^{S}$
for edge $w_{i}^{S}\xrightarrow{SBV}w_{j}^{S}$ in the dependency
tree). Then for each pair of words $(w_{i}^{S},w_{j}^{S})$, the
shortest path with labels $L=[w_{1}^{L},\cdots,w_{N_{L}}^{L}]$ in
the graph is computed and mapped into a sequence of $N_{e}$-dimensional
distributed representation vectors $[l_{1},\cdots,l_{N_{L}}]$ by the
embedding operation. One can employ RNN network to convert this sequence
of vectors into a feature vector, but RNN operation is time-consuming.
We simply concatenate vectors in short paths ($N_{L}\le$3) into a
$3N_{e}$ dimensional vector and feed the vector into a two-layer
feed forward neural network to generate an $N_{h}$-dimensional feature
vector $n_{ij}$. For long paths with $N_{L}>3$, $n_{ij}$ is set
to a zero vector. We define dependency attention vector $\tilde{u}_{j}^{t}=\sum_{i}p^{*}(w_{t}^{F}=w_{i}^{S})n_{ij}$,
where $p^{*}$ is the sharpened probability $p$ defined in Equation~(\ref{eq:total_prob}).
If $w_{t}^{F}\in S$, $\tilde{u}_{j}^{t}$ represents the semantic
relationship between $w_{t}^{F}$ and $w_{j}^{S}$. If $w_{t}^{F}\in K$,
then $\tilde{u}_{j}^{t}$ is close to zero. To correctly guide the
decoder, we need to gate  $\tilde{u}_{j}^{t}$ to remember the previous
attention vector sometimes (for example, when $\$$ is selected),
and to forget it sometimes (for example, when a new fact is started).
Finally, we define $u_{j}^{t}=g($$\tilde{u}_{j}^{t}$) as the gated
dependency attention vector, where $g$ is the GRU gated function,
and update the alignment model as follows:
\begin{eqnarray*}
e_{tj} & = & a(s_{t-1},h_{j}^{S},m_{j}^{t-1},u_{j}^{t-1})\\
 & = & v_{a}^{T}\tanh(W_{a}s_{t-1}+U_{a}h_{j}+V_{a}m_{j}^{t-1}+D_{a}u_{j}^{t-1})
\end{eqnarray*}
where $D_{a}\in\mathbb{R}^{(N_{h}\times N_{h})}$.

\subsection{Post processing}

For each sequence generated by Logician, we parse it into a set
of facts, remove tuples with illegal format or duplicated tuples. The resultant
set is taken as the output of the Logician.

\section{Empirical Evaluation\label{sec:Experimental-Results}}

\subsection{Experimental Design }

We first measure the utility of various components in Logician
to select the optimal model, and then compare this model to the state-of-the-art
methods in four types of information extraction tasks: verb/preposition-based
relation, nominal attribute, descriptive phrase and hyponymy relation.
The SAOKE data set is split into training set, validating set and
testing set with ratios of 80\%, 10\%, 10\%, respectively. For all
algorithms involved in the experiments, the training set can be used
to train the model, the validating set can be used to select an optimal
model, and the testing set is used to evaluate the performance.

\subsubsection{Evaluation Metrics\label{subsec:Evaluation-Metrics}}

For each instance pair $(S,F)$ in the test set, where $S$ is the
input sentence and $F$ is the formatted string of ground truth of
facts, we parse $F$ into a set of tuples $\mathbb{F}=\{F_{i}\}_{j=1}^{M}$.
Given an open information extraction algorithm, it reads $S$ and
produces a set of tuples $\mathbb{G}=\{G_{i}\}_{j=1}^{N}$. To evaluate
how well the $\mathbb{G}$ approximates $\mathbb{F}$, we need to
match each $G_{i}$ to a ground truth fact $F_{j}$ and check whether
$G_{i}$ tells the same fact as $F_{j}$. To conduct the match, we compute
the similarity between each predicted fact in $\mathbb{G}$ and each
ground truth fact in $\mathbb{F}$, then find the optimal matching
to maximize the sum of matched similarities by solving a linear assignment
problem~\cite{assignment_problem}. In the procedure, the similarity
between two facts is defined as
\[
Sim(G_{i},F_{j})=\frac{\sum_{l=1}^{\min(\mathbf{n}(G_{i}),\mathbf{n}(F_{j}))}\mathbf{g}(G_{i}(l),F_{j}(l))}{\max(\mathbf{n}(G_{i}),\mathbf{n}(F_{j}))},
\]
where $G_{i}(l)$ and $F_{j}(l)$ denote the $l$-th element of tuple
$G_{i}$ and $F_{j}$ respectively, $\mathbf{g}(\cdot,\cdot)$ denotes
the gestalt pattern matching~\cite{Ratcliff:1988:PMG} measure for
two strings and $\mathbf{n}(\text{\ensuremath{\cdot})}$ returns the
length of the tuple.

Given a matched pair of $G_{i}$ and $F_{j}$ , we propose an automatic
approach to judge whether they tell the same fact. They are judged
as telling the same fact if one of the following two conditions
is satisfied:
\begin{itemize}
\item $\mathbf{n}(G_{i})=\mathbf{n}(F_{j})$, and $\mathbf{g}(G_{i}(l),F_{j}(l))\ge0.85,l=1,\cdots,\mathbf{n}(G_{i})$;
\item $\mathbf{n}(G_{i})=\mathbf{n}(F_{j})$, and $\mathbf{g}(\mathcal{S}(G_{i}),\mathcal{S}(F_{j})\ge0.85$;
\end{itemize}
where $\mathcal{S}$ is a function formatting a fact into a string
by filling the arguments into the placeholders of the predicate.

With the automatic judgment, the precision ($P$), recall ($R$)
and $F_{1}$-score over a test set can be computed. By defining a
confidence measure and ordering the facts by their confidences, a
precision-recall curve can be drawn to illustrate the overall performance
of the algorithm. For Logician, the confidence of a fact is computed
as the average of log probabilities over all words in that fact.

Beyond the automatic judgment, human evaluation is also employed.
Given an algorithm and the corresponding fact confidence measure,
we find a threshold that produces approximately 10\% recall (measured
by automatic judgment) on the validation set of SAOKE data set. A
certain number of sentences (200 for verb/preposition based relation
extraction task, and 1000 for other three tasks) are randomly chosen
from the testing set of SAOKE data set, and the facts extracted from
these sentences are filtered with that threshold. Then we invite three
volunteers to manually refine the labeled set of facts for each sentence
and vote to decide whether each filtered fact is correctly involved
in the sentence. The standard precision, recall and $F_{1}$-score are reported
as the human evaluation results.

\subsubsection{Training the Logician Model}

For each instance pair $(S,F)$ in the training set of SAOKE data
set, we split $S$ and $F$ into words using LTP toolset~\cite{Che2010},
and words appearing in more than 2 sentences are added to the vocabulary.
By adding the OOV (out of vocabulary) tag, we finally obtain a vocabulary
$V$ with size $N_{V}=65,293$. The dimension of all embedding vectors
is set to $N_{e}=200$, and the dimension of hidden states is set
to $N_{h}=256$. We use a three-layer bi-directional GRU with dimension
128 to encode $\{x_{i}\}_{i=1}^{N_{S}}$ into hidden states $\{h_{i}^{S}\}_{i=1}^{N_{S}}$,
and a two-layer GRU with hidden-dimension 256 to encode the sequence
of $\{y_{j}\}_{j=1}^{N_{F}}$ into hidden states $\{h_{j}^{F}\}_{j=1}^{N_{F}}$.
Finally, the Logician network is constructed as stated in Section~\ref{sec:Logician}.
 The Logician is then trained using stochastic gradient descent
(SGD) with RMSPROP~\cite{Hinton2012x} strategy for 20 epochs with
batch size 10 on the training set of SAOKE data set. The model with
best $F_{1}$-score by automatic judgment on the validation set is
selected as the trained model. When the model is trained,
given a sentence, we employ the greedy search procedure to produce the fact sequences.

\subsection{Evaluating Components' Utilities}

In this section, we analyze the effects of components involved in
Logician: restricted copy, coverage, and gated dependency. Since
the restricted copy mechanism is the essential requirement of Logician
in order to achieve the goal of literally honest, we take the Logician with
only copy mechanism (denoted by $Copy$) as the baseline, and analyze
the effeteness of coverage mechanism (denoted by $Copy+Coverage$),
gated dependency mechanism (denoted by $Copy+GatedDep$) and both
(denoted by $All$). Furthermore, there is another option of whether
or not to involve shallow semantic information such as POS-tag and
NER-tag into the model. For models involving such information, the
POS-tag and NER-tag of each word in sentence $S$ are annotated using
LTP. For each word in $F$ that is not any keyword in $K$, the POS-tag
and NER-tag are copied from the corresponding original word in $S$.
For each keyword in $K$, a unique POS-tag and a unique NER-tag
are assigned to it. Finally, for each word in $S$ or $F$, the POS-tag
and NER-tag are mapped into $N_{e}$-dimensional distributed representation
vectors and are concatenated into $x_{i}$ or $y_{j}$ to attend the
training.

All models are trained using the same settings described in above
section, and the default output facts (without any confidence filtering)
are evaluated by the automatic judgment. The results are reported
in Table~\ref{tab:Analysis-of-Components}. From the results, we
can see that the model involving all the components and shallow tag
information archives the best performance. We use that model to attend
the comparisons with existing approaches.

\begin{table}
\begin{centering}
\caption{Analysis of Components involved in Logician.\vspace{-0in}\label{tab:Analysis-of-Components}}
\par\end{centering}
\begin{centering}
\begin{tabular}{cccc|ccc}\hline
 & \multicolumn{3}{c|}{With Shallow Tag} & \multicolumn{3}{c}{No Shallow Tag}\tabularnewline
\cline{2-7}
 & $P$ & $R$ & $F_{1}$ & $P$ & $R$ & $F_{1}$\tabularnewline
\hline
\hline
$Copy$ & 0.375 & 0.412 & 0.393 & 0.378 & 0.431 & 0.403\tabularnewline
\hline
$Copy+Coverage$ & 0.390 & 0.390 & 0.390 & 0.307 & 0.417 & 0.353\tabularnewline
\hline
$Copy+GatedDep$ & 0.427 & 0.371 & 0.397 & 0.335  & 0.419 & 0.372\tabularnewline
\hline
$All$ & 0.443 & 0.419 & \textbf{0.431} & 0.373 & 0.435 & 0.402\tabularnewline
\hline
\end{tabular}
\par\end{centering}
\end{table}

\subsection{Comparison with Existing Approaches\label{subsec:Comparison-with-Existing}}

\subsubsection{Verb/preposition-Based Relation Extraction }

In the task of extracting verb/preposition based facts, we compare
our Logician with the following state-of-the-art Chinese OIE algorithms:

\textbf{\noun{SRLIE}}: our implementation of SRLIE~\textbf{\noun{\cite{Christensen2011}}}
for the Chinese language, which first uses LTP tool set to extract
the semantic role labels, and converts the results into fact tuples
using heuristic rules. The confidence of each fact is computed as
the ratio of the number of words in the fact to the number of words
in the shortest fragment of source sentence that contains all words
in the fact.

\textbf{\noun{ZORE}} : the Chinese Open Relation Extraction system~\cite{Qiu2014},
which builds a set of patterns by bootstrapping based on dependency
parsing results, and uses the patterns to extract relations. We used
the program provided by the author of ZORE system ~\cite{Qiu2014}
to generate the extraction results in XML format, and developed an
algorithm to transform the facts into n-ary tuples, where auxiliary information extracted by ZORE is removed. The confidence
measure for ZORE is the same as that for SRLIE.

\textbf{\noun{SRL$_{\text{SAOKE}}$}}: our implementation of the states-of-the-art
SRL algorithm proposed in~\cite{He2017} with modifications to fit
OIE tasks.
$\text{SRL}_{\text{SAOKE}}$ extracts facts in two steps:
(i) Predicate head word detection: detects head word for predicate
of each possible fact, where head word of a predicate is the last
word in the predicate depending on words outside the predicate in
the dependency tree. (ii) Element phrase detection: For each detected
head word, detects the subject phrase, predicate phrase and object
phrases by tagging the sentence with an extended BIOE tagging scheme,
which tags the word neighboring the separation point of the phrase by ``M'' to cope with the separated phrase. We modify the code provided by the author of~\cite{He2017}
to implement above strategy, and then train a model with the same
parameter setting in~\cite{He2017} on the training set of SAOKE
data set. The confidence measure for $\text{SRL}_{\text{SAOKE}}$
is computed as the average of log probabilities over all tags of words
in facts. Note that $\text{SRL}_{\text{SAOKE}}$ can extract both
verb/preposition based relation and nominal attributes, but in this
section, we only evaluate the results of the former type of facts.

\begin{figure}

\mbox{\hspace{-0.3in}
\subfloat[Verb/preposition-based Relation\label{fig:Verb/preposition-based-Relation}]
{\includegraphics[width=0.6\columnwidth]{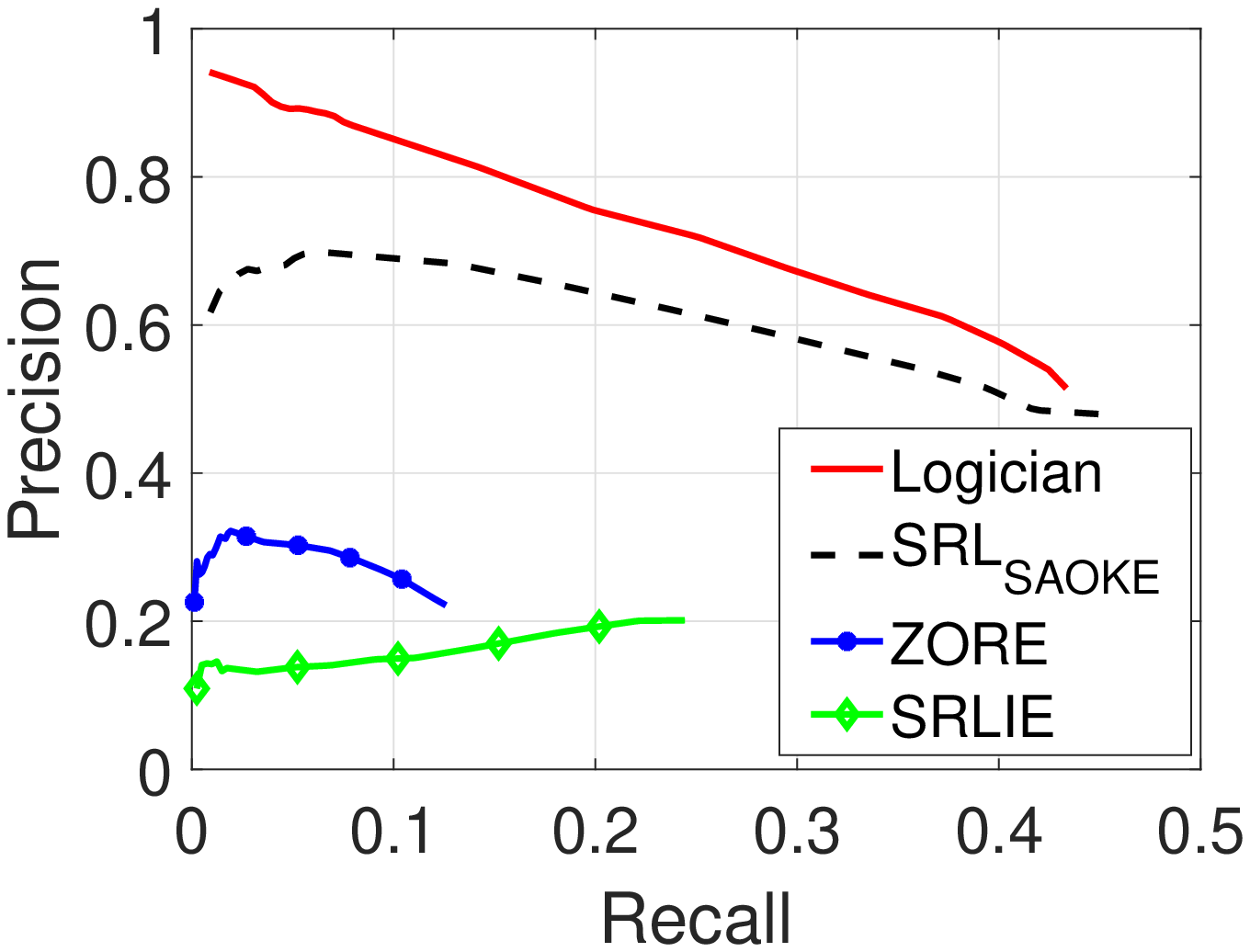}}\hspace{-0.2in}
\subfloat[Nominal Attribute\label{fig:Nominal-Attribute}]{
\includegraphics[width=0.6\columnwidth]{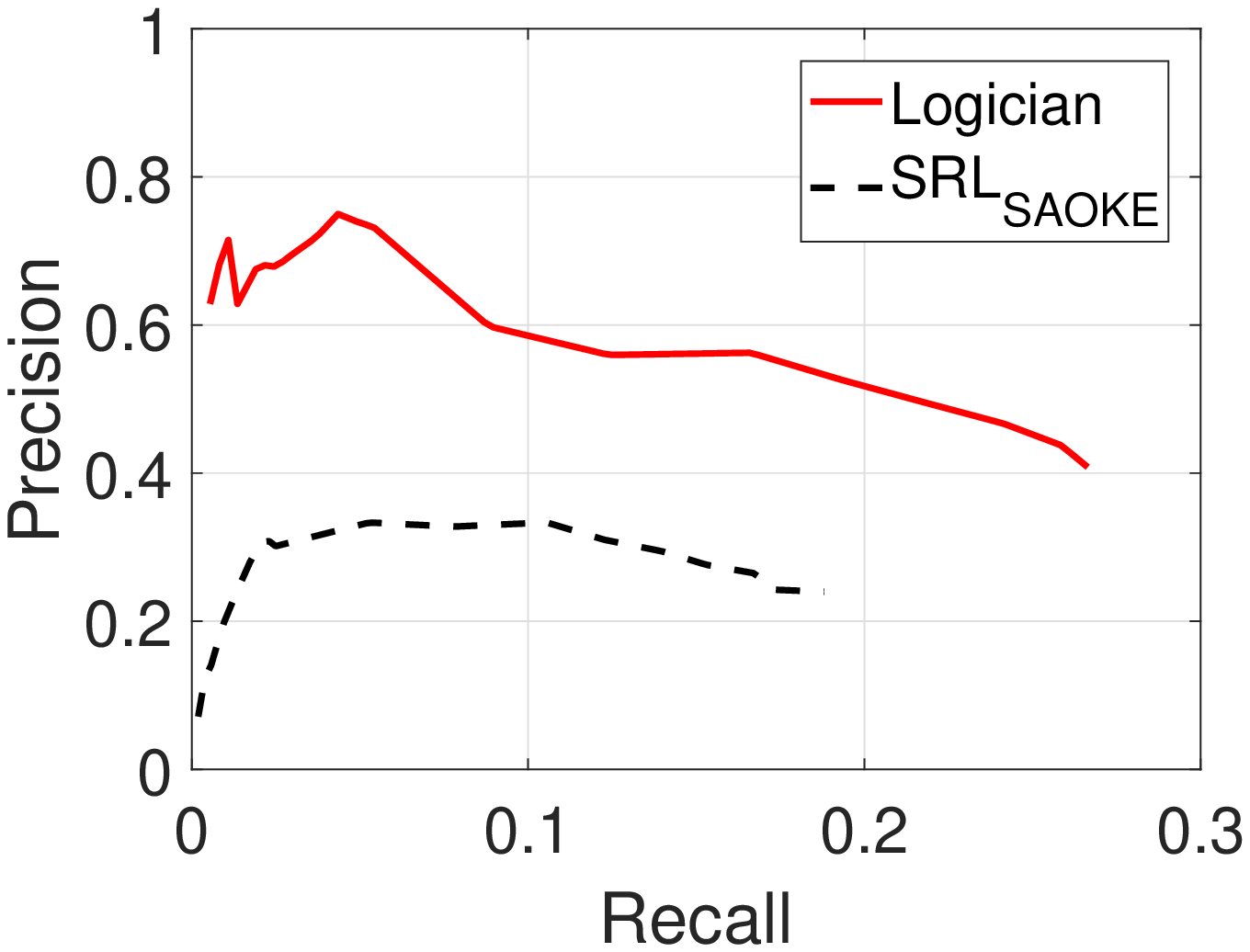}}
}
\mbox{\hspace{-0.3in}
\subfloat[Descriptive Phrase\label{fig:Descriptive-Phrase}]{
\includegraphics[width=0.6\columnwidth]{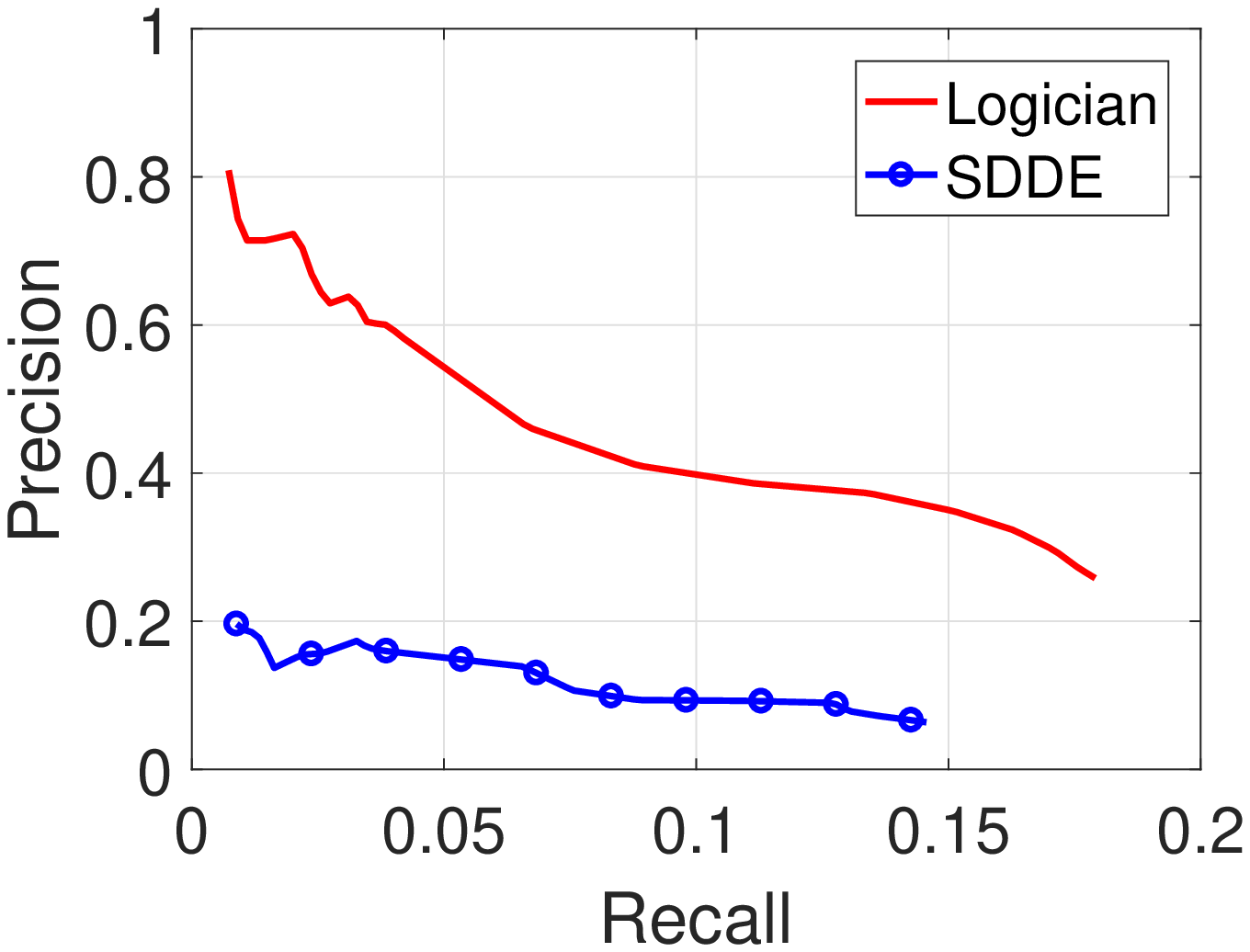}} \hspace{-0.2in}
\subfloat[Hyponymy\label{fig:Hyponymy}]{
\includegraphics[width=0.6\columnwidth]{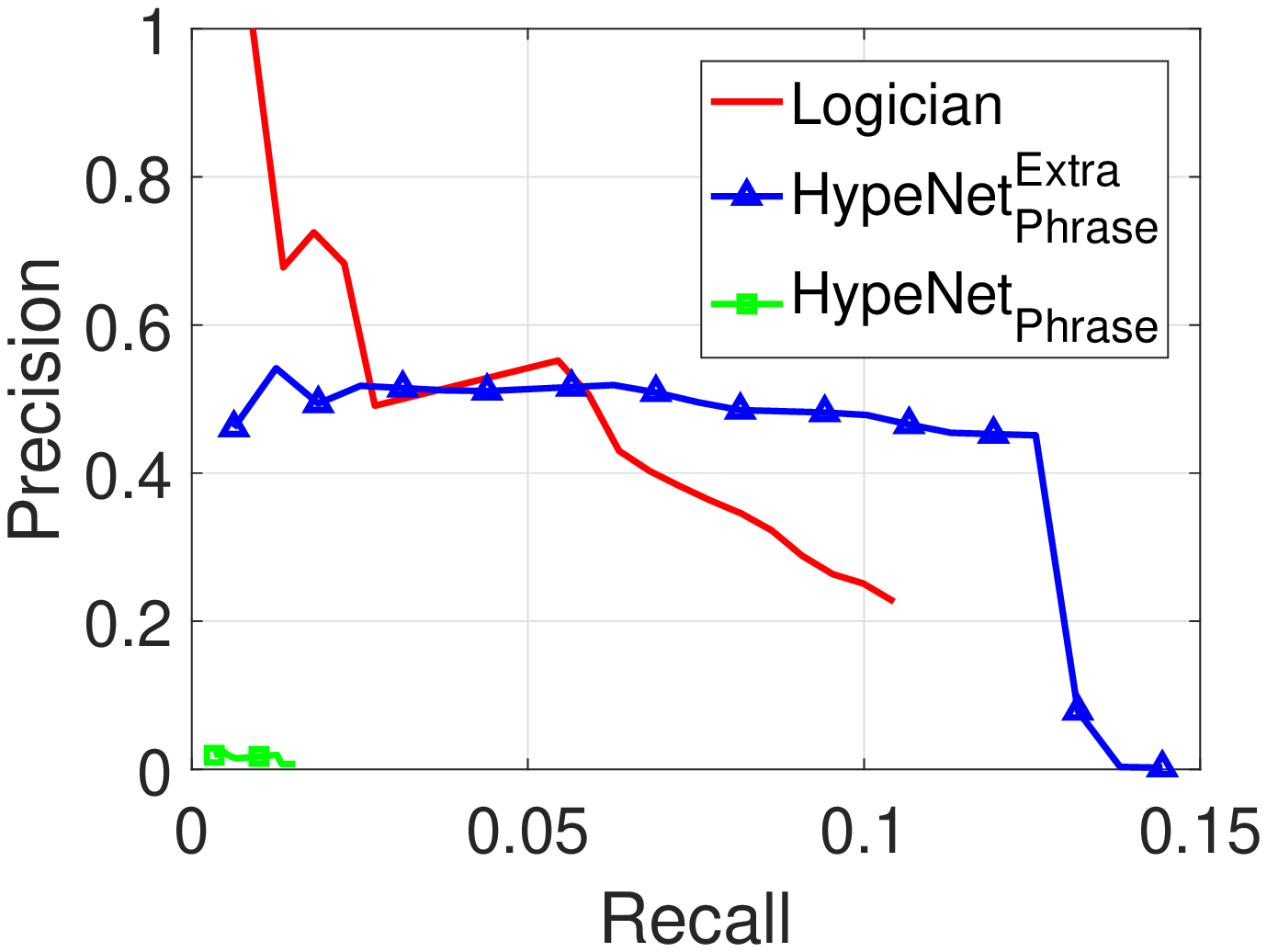}
}
}
\caption{Performance comparison on four types of information extraction tasks.}
\end{figure}

The precision-recall curves of Logician and above three comparison
algorithms are shown in Figure~\ref{fig:Verb/preposition-based-Relation},
and the human evaluation results are shown in the first section of
Table~\ref{tab:open-fact-results}.

\subsubsection{Nominal Attribute Extraction}

The state-of-the-art\\ nominal attribute extraction method is ReNoun~\cite{Pal2016,Yahya2014}.
However, it relies on a pre-constructed English attribute schema system~\cite{Gupta2014}
which is not available for Chinese, so it is not an available baseline for Chinese. Since $\text{SRL}_{\text{SAOKE}}$ can extract nominal
attributes, we compare Logician with $\text{SRL}_{\text{SAOKE}}$
on this task.  The precision-recall curves of Logician and $\text{SRL}_{\text{SAOKE}}$
on the nominal attribute extraction task are shown in Figure~\ref{fig:Nominal-Attribute},
and the human evaluation results are shown in the second section of
Table~\ref{tab:open-fact-results}.

\subsubsection{Descriptive Phrase Extraction}

Descriptive phrase extraction has been considered in~\cite{Chakrabarti2011},
in which domain names are required to develop patterns to extract
candidates for descriptive phrases, so this method is not applicable
to open domain tasks. We develop a baseline
algorithm (called Semantic Dependency Description Extractor, SDDE)
to extract descriptive phrase. It extracts
semantic dependency relation between words using LTP toolset, and
for each noun $w_n$ which is the parent of some semantic ``Desc'' relations, identifies
a noun phrase $N$ with $w_n$ as its heading word,
assembles a descriptive phrase $D$ containing all words with ``Desc''
relation to $w_n$, and finally outputs the fact ``($N$, $DESC$, $D$)''. The confidence of fact in SDDE is computed as the ratio of
the number of adverbs and adjectives in $D$ to the number of words in $D$. The precision-recall curves of Logician and SDDE on the
descriptive phrase extraction task are shown in Figure~\ref{fig:Descriptive-Phrase},
and the human evaluation results are shown in the third section of
Table~\ref{tab:open-fact-results}.

\subsubsection{Hyponymy Extraction}

HypeNet~\cite{VeredShwartz2016a} is the state-of-the-art
algorithm recommended for hyponymy extraction~\cite{Wang2017a}, which judges
whether hyponymy relation exists between two given words. To make
it capable of judging hyponymy relation between two phrases, we replace the word embedding vector component in HypeNet by an LSTM network. Two modified HypeNet models are built using different training data sets: (i) $\text{HypeNet}_{\text{Phrase}}$: using the pairs of phrases with ISA relation
in the training set of SAOKE data set (9,407 pairs after the compact
expression expansion); (ii) $\text{HypeNet}_{\text{Phrase}}^{\text{Extra}}$: besides the training set for $\text{HypeNet}_{\text{Phrase}}$, adding two Chinese hyponymy data sets (1.4 million pair of words in total in hyponymy relation): Tongyici Cilin (Extended)
(CilinE for short)~\cite{Che2010} and cleaned Wikipedia Category
data~\cite{Li2015}. In both cases,
the sentences from both Chinese Wikipedia pages and training set of SAOKE data set are taken as the background corpus for the HypeNet algorithm. In the testing phase, the trained models are used to predict whether the hyponymy relation exists for each pair of noun phrases/words in sentences of the testing set of SAOKE data set. The confidence of a judgment is the predicted probability of the existence of hyponymy relation.
The precision-recall curves of Logician, $\text{HypeNet}_{\text{Phrase}}$
and $\text{HypeNet}_{\text{Phrase}}^{\text{Extra}}$ are shown in Figure~\ref{fig:Hyponymy},
and the human evaluation results in the fourth section of
Table~\ref{tab:open-fact-results}.

\begin{table}[h!]
\caption{Human evaluation results on four types of information extraction tasks.\vspace{-0in}\label{tab:open-fact-results}}
\begin{centering}
\begin{tabular}{lcccc}
\hline
 Task &Method  & $P$ & $R$ & $F_{1}$\tabularnewline
\hline
\hline
\multirow{4}{*}{1.\ \ Relation} & SRLIE & 0.166  & 0.119 & 0.139 \tabularnewline
\cline{2-5}
 & ZORE & 0.300  & 0.136 & 0.187 \tabularnewline
\cline{2-5}
 & $\text{SRL}_{\text{SAOKE}}$  & 0.610  & 0.130 & 0.214 \tabularnewline
\cline{2-5}
 & Logician & 0.883 & 0.147 & \textbf{0.252} \tabularnewline
\hline
\multirow{2}{*}{2.\ \ Attribute} & $\text{SRL}_{\text{SAOKE}}$  & 0.362 & 0.102 & 0.159 \tabularnewline
\cline{2-5}
 & Logician & 0.603 & 0.126 & \textbf{0.208} \tabularnewline
\hline
\multirow{2}{*}{3.\ \ Description} & SDDE & 0.135 & 0.146 & 0.140\tabularnewline
\cline{2-5}
 & Logician & 0.392 & 0.109 & 0.\textbf{170} \tabularnewline
\hline
\multirow{3}{*}{4.\ \ Hyponymy} & $\text{HypeNet}_{\text{Phrase}}$ & 0.007 & 0.016 & 0.010\tabularnewline
\cline{2-5}
 & $\text{HypeNet}_{\text{Phrase}}^{\text{Extra}}$ & 0.442 & 0.090 & 0.149\tabularnewline
\cline{2-5}
 & Logician & 0.317 & 0.129 & \textbf{0.183}\tabularnewline
\hline
\end{tabular}
\par\end{centering}
\end{table}

\vspace{-2mm}

\subsection{Results Analysis}

The experimental results reveal that, Logician outperforms the
comparison methods with large margin in first three tasks. For hyponymy
detection tasks, Logician overwhelms the $\text{HypeNet}_{\text{Phrase}}$
using the same training data, and produces comparable results to $\text{HypeNet}_{\text{Phrase}}^{\text{Extra}}$ with much less training data. Table~\ref{tab:openie-Case-Study} exhibits several
example sentences and the facts extracted by these algorithms.

The poor performance of pattern-based methods is plausibly due to the noise
in SAOKE data set. The sentences in SAOKE data set are randomly selected
from a web encyclopedia, with free and casual writing style,
are thus more noisy than the training data of NLP toolset used by these
methods. In this situation, the NLP toolset may produce poor results,
so do the pattern-based methods.

Models learned from the SAOKE data set archive much better performance. Nevertheless, $\text{SRL}_{\text{SAOKE}}$
extracts each fact without knowing whether a candidate word has been
used in other facts, which results in the misleading overlap of the
word \begin{CJK}{UTF8}{gbsn}``学''\end{CJK} (``Learn'' in English)
between two facts in the first case of Table~\ref{tab:openie-Case-Study}.
Similarly, $\text{HypeNet}_{\text{Phrase}}$ and $\text{HypeNet}_{\text{Phrase}}^{\text{Extra}}$
focus on the semantic vectors of pairs of phrases and their dependency
paths in the background corpus. They extract each fact independently from other facts and hence  do not know whether there have been any other relations extracted about these two phrases. In other words, for those comparison methods, an important source of information is neglected and a global optimization for all facts involved in sentences is absent.

On the contrary, Logician performs
global optimization over the facts involved in each sentence by the sequence-to-sequence
learning paradigm with the help of the coverage mechanism, in which
facts compete each other to attract the attention of words, but also
cooperate to share words. Valuable information is shared between these
multiple tasks, which makes Logician consistently superior to other
algorithms in these tasks.

Furthermore, $\text{SRL}_{\text{SAOKE}}$
and $\text{HypeNet}$ methods suffer from the OOV problem, such as
unfamiliar words/phrases like the person name and school name in the
last case of Table~\ref{tab:openie-Case-Study}. In this situation
they may fail to produce a reasonable result. Logician is able
to cope with unfamiliar words/phrases by exploiting the context information
using deep RNN network with the help of copy mechanism.

\begin{table*}
\begin{centering}
\caption{Example of extraction from a sentence for each task.\vspace{-0in}\label{tab:openie-Case-Study}}
\par\end{centering}

\small
\begin{centering}
\begin{tabular}{>{\raggedright}m{0.10\textwidth}>{\centering}m{0.08\textwidth}>{\raggedright}m{0.28\textwidth}>{\raggedright}m{0.42\textwidth}>{\centering}m{0.025\textwidth}}
\hline
 &  & Chinese  & English Translation & Label\tabularnewline
\hline
\hline
 & Sentence & \begin{CJK}{UTF8}{gbsn}学道访仙，晚年修道于石笋山\end{CJK} & Learn the Tao, visit the immortal, practice Taoism in the Stalagmite
Hill in the old age & \tabularnewline
\cline{2-5}
 & SRLIE & \begin{CJK}{UTF8}{gbsn}(X,学,道)(X,访,仙)   \end{CJK}  & \textcolor{black}{(X, learn, the Tao)}(X, visit, the immortal) & \cmark\tabularnewline
\cline{2-5}
1.\ \ Relation & ZORE & \begin{CJK}{UTF8}{gbsn}(学道,访,仙)\newline(学道,修道于,石笋山 )\end{CJK} & \textcolor{black}{(Learn the Tao, visit, the immortal)}

(Learn the Tao, practice Taoism in, the Stalagmite Hill) & \xmark

\xmark\tabularnewline
\cline{2-5}
 & $\text{SRL}_{\text{SAOKE}}$ & \begin{CJK}{UTF8}{gbsn}(X,学,道访)(学,修道于,石笋山) \end{CJK} & \textcolor{black}{(X, learn, Tao visit)}(Learn, practice Taoism in,
the Stalagmite Hill) & \xmark\tabularnewline
\cline{2-5}
 & Logician & \begin{CJK}{UTF8}{gbsn}(X,学,道)(X,访,仙)\newline(X,修道于,石笋山) \end{CJK} & \textcolor{black}{(X, learn, the Tao)}(X, visit, the immortal)

(X, practice Taoism in, the Stalagmite Hill) & \cmark

\cmark\tabularnewline
\hline
\hline
 & Sentence & \begin{CJK}{UTF8}{gbsn}全村辖区总面积约10平方公里，其中耕地132公顷。\end{CJK} & The whole village area of about 10 square kilometers, of which 132
hectares of arable land. & \tabularnewline
\cline{2-5}
2.\ \ Attribute & $\text{SRL}_{\text{SAOKE}}$ & \begin{CJK}{UTF8}{gbsn}(全村辖区,总面积,约10平方公里)\end{CJK} & (The village, whole area, about 10 square kilometers) & \cmark\tabularnewline
\cline{2-5}
 & Logician & \begin{CJK}{UTF8}{gbsn}(全村辖区,总面积,约10平方公里)\newline(全村,耕地,132公顷)\end{CJK} & (The village, whole area, about 10 square kilometers)

(The village, arable land, 132 hectares) & \cmark

\cmark\tabularnewline
\hline
\hline
 & Sentence & \begin{CJK}{UTF8}{gbsn}硫酸钙较为常用，其性质稳定，无嗅无味，微溶于水\end{CJK} & Calcium sulfate is commonly used, its properties are stable, odorless
and tasteless, slightly soluble in water & \tabularnewline
\cline{2-5}
3.\ \ Description & SDDE & No Recall & No Recall & \tabularnewline
\cline{2-5}
\multirow{1}{0.08\textwidth}{} & Logician & \begin{CJK}{UTF8}{gbsn}(硫酸钙,DESC,[性质稳定|无嗅无味])\newline(硫酸钙,DESC,稳定)\end{CJK} & \textcolor{black}{(Calcium sulfate, DESC, {[}properties stable | odorless
and tasteless{]})}

\textcolor{black}{(Calcium sulfate, DESC, stable)} & \cmark

\xmark\tabularnewline
\hline
\hline
 & Sentence & \begin{CJK}{UTF8}{gbsn}蔡竞，男，汉族，四川射洪人，西南财经大学经济学院毕业，经济学博士。 \end{CJK} & \textcolor{black}{Cai Jing, male, Han Chinese, a Sichuan Shehong native,
graduated from the economics school of Southwestern University of
Finance and Economics (SUFE), and a Ph. D. in economics.} & \tabularnewline
\cline{2-5}
4.\ \ Hyponymy & $\text{HypeNet}_{\text{Phrase}}$ & \begin{CJK}{UTF8}{gbsn}
(经济学博士,ISA, 蔡)
\end{CJK} & \textcolor{black}{(Ph. D. in economics,ISA,Cai)} & \xmark\tabularnewline
\cline{2-5}
 & $\text{HypeNet}_{\text{Phrase}}^{\text{Extra}}$ & \begin{CJK}{UTF8}{gbsn}
(西南财经大学经济学院,ISA,经济学博士)\newline (西南财经大学经济学院,ISA,四川)\newline (经济学博士,ISA,西南财经大学经济学院)
\end{CJK} & \textcolor{black}{(the economics school of SUFE, ISA, Ph. D. in economics) }

\textcolor{black}{(the economics school of SUFE,ISA,Sichuan)}

\textcolor{black}{(Ph. D. in economics,ISA,the economics school of
SUFE)} & \xmark

\xmark

\xmark\tabularnewline
\cline{2-5}
 & Logician & \begin{CJK}{UTF8}{gbsn}
(蔡竞,ISA,[男|汉族|四川射洪人])
\end{CJK} & (Cai Jing,ISA,{[}male |Han Chinese | Sichuan Shehong native{]}) & \cmark\tabularnewline
\hline
\end{tabular}
\par\end{centering}
\vspace{-3mm}
\centering{}
\end{table*}

\subsection{Extraction Error Analysis of Logician}

We do a preliminary analysis for the results produced by the Logician
model. The most notable problem is that it is unable to recall some facts for
long or complex sentences. The last case in Table~\ref{tab:openie-Case-Study}
exhibits such situation, where the fact \begin{CJK}{UTF8}{gbsn}(蔡竞,ISA,经济学博士)\end{CJK}((Cai
Jing, ISA, \textcolor{black}{Ph. D. in economics}) in English) is
not recalled. This phenomenon indicates that the coverage mechanism
may lose effectiveness in this situation. The second class of error
is incomplete extraction, as exhibited in the third case in Table~\ref{tab:openie-Case-Study}.
Due to the incomplete extraction, the left parts may interfere the
generation of other facts, and result in nonsense results, which is
the third class of error. We believe it is helpful to introduce extra
rewards into the learning procedure of Logician to overcome these
problems. For example, the reward could be the amount of remaining
information left after the fact extraction, or the completeness of
extracted facts. Developing such rewards and reinforcement learning
algorithms using those rewards to refine Logician belongs to our
future works.

\section{Related Works\label{sec:Related-work}}

\subsection{Knowledge Expressions}

\label{subsec:knowledge_express}

Tuple is the most common knowledge expression format for OIE systems
to express n-ary relation between subject and objects. Beyond such
information, ClausIE~\cite{Corro2013} extracts extra information
in the tuples: a complement, and one or more adverbials, and OLLIE~\cite{Schmitz2012}
extracts additional context information. SAOKE is able to express
n-ary relations, and can be easily extended to support the knowledge
extracted by ClausIE, but needs to be redesigned to support context
information, which belongs to our future work.

However, there is a fundamental difference between SAOKE and tuples
in traditional OIE systems. In traditional OIE systems, knowledge
expression is generally not directly related to the extraction algorithm.
It is a tool to reorganize the extracted knowledge into a form for
further easy reading/storing/computing. However, SAOKE is proposed
to act as the direct learning target of the end-to-end Logician
model. In such end-to-end framework, knowledge representation is the
core of the system, which decides what information would be extracted
and how complex the learning algorithm would be. To our knowledge,
SAOKE is the first attempt to design a knowledge expression friendly
to the end-to-end learning algorithm for OIE tasks. Efforts are still needed
to make SAOKE more powerful in order to express more complex knowledge
such as events.

\subsection{Relation Extraction}

Relation extraction is the task to identify semantic connections between
entities. Major existing relation extraction algorithms can be classified
into two classes: closed-domain and open-domain. Closed-domain algorithms
are learnt to identify a fixed and finite set of relations, using
supervised methods~\cite{Kambhatla2004,Zelenko2003,Miwa2016,Zheng2017}
or weakly supervised methods~\cite{Mintz2009,Lin2016}, while the
open-domain algorithms, represented by aforementioned OIE systems,
discover open-domain relations without predefined schema. Beyond these
two classes, methods like universal schema~\cite{Riedel2013a} are
able to learn from both data with fixed and finite set of relations,
such as relations in Freebase, and data with open-domain surface relations
produced by heuristic patterns or OIE systems.

Logician can be used as an OIE system to extract open-domain relation
between entities, and act as sub-systems for knowledge base construction/completion
with the help of schema mapping~\cite{Soderland2010}. Compared with
existing OIE systems, which are pattern-based or self-supervised by
labeling samples using patterns~\cite{Mausam2016}, to our knowledge
Logician is the first model trained in a supervised end-to-end
approach for OIE task, which has exhibited powerful ability in our
experiments. There are some neural based end-to-end systems~\cite{Miwa2016,Zheng2017,Lin2016}
proposed for relation extraction, but they all aim to solve the close-domain
problem.

However, Logician is not limited to relation extraction task. First,
Logician extracts more information beyond relations. Second, Logician
focuses on examining how natural languages express facts~\cite{Etzioni2011},
and producing helpful intermediate structures for high level tasks.

\subsection{Language to Logic }

Efforts had been made to map natural language sentences into logical
form. Some approaches such as~\cite{Kate2005,Zettlemoyer2005,Yin2015,Dong2016a}
learn the mapping under the supervision of manually labeled logical
forms, while others~\cite{Cai2013b,Kwiatkowski2013} are indirectly
supervised by distant information, system rewards, etc. However, all
previous works rely on a pre-defined, domain specific logical system, which
limits their ability to learn facts out of the pre-defined logical
system.

Logician can be viewed as a system that maps language to \emph{natural
logic}, in which the majority of information is expressed by natural
phrase. Other than systems mentioned above which aim at execution
using the logical form, Logician focuses on understanding how the
fact and logic are expressed by natural language. Further mapping
to domain-specific logical system or even executor can be built on
the basis of Logician's output, and we believe that, with the help
of Logician, the work would be easier and the overall performance of the system may be improved.
 
 \subsection{Facts to Language}

The problem of generating sentences from a set of facts has  attracted a lot of attentions ~\cite{Wiseman2017,Chisholm2017,Agarwal2017,Vougiouklis2017}. These models focus on facts with a predefined schema from a specific problem domain, such as people biographies and basketball game records, but could not work on open domain. The SAOKE data set provides an opportunity to extend the ability of these models into open domain. 
 
\subsection{Duality between Knowledge and Language}

As mentioned in above sections, the SAOKE data set provides examples of dual mapping between facts and sentences. Duality has been verified to be useful to promote the performance of agents in many NLP tasks, such as back-and-forth translation~\cite{Xia2016a}, and question-answering~\cite{tang2017question}. It is a promising approach to use the duality between knowledge and language to improve the performance of Logician. 

\section{Conclusion\label{sec:Conclusion}}

In this paper, we consider the open information extraction (OIE) problem
for a variety of types of facts in a unified view.  Our solution consists of three components: \textbf{SAOKE format}, \textbf{SAOKE data set}, and \textbf{Logician}.  SAOKE form is designed to express different types of facts in a unified manner. We publicly
release the largest manually labeled data set for OIE tasks in SAOKE
form. Using the labeled SAOKE data set, we train an end-to-end neural
sequence-to-sequence model, called Logician, to transform sentences
in natural language into facts. The experiments reveal the superiority
of Logician in various open-domain information extraction
tasks to the state-of-the-art algorithms.

Regarding future work, there are at least three promising directions.
Firstly, one can investigate knowledge expression methods to extend SAOKE to
express more complex knowledge, for tasks such as event extraction.
Secondly, one can develop novel learning strategies to improve the performance
of Logician and adapt the algorithm to the extended future version of
SAOKE. Thirdly, one can extend SAOKE format and Logician algorithm in other languages.

\balance
\bibliographystyle{plain}
\bibliography{logician}

\end{document}